\documentclass[letterpaper, 10 pt, conference]{IEEEtran}
\IEEEoverridecommandlockouts

\usepackage[numbers]{natbib}
\usepackage{multicol}
\usepackage{times}
\usepackage[bookmarks=true]{hyperref}

\usepackage{graphics}
\usepackage{amsmath}
\usepackage{amssymb}  
\usepackage{siunitx}
\usepackage{mathtools}
\usepackage{booktabs}
\usepackage[ruled]{algorithm2e}
\usepackage{caption}
\usepackage{subcaption}
\usepackage[bottom=57pt,top=54pt, left=54pt, right=54pt]{geometry}   
\usepackage{adjustbox}
\usepackage{pbox}
\usepackage{dblfloatfix}
\usepackage{graphicx}
\usepackage[dvipsnames]{xcolor}

\usepackage{array}
\usepackage{multirow}
\usepackage{makecell}
\usepackage{pifont}
\usepackage{xparse}
\usepackage{enumitem}
\usepackage{float}
\floatstyle{plaintop}
\restylefloat{table}
\usepackage{arydshln}
\usepackage{bm}
\usepackage{soul}

\NewDocumentCommand{\todo}{o m}{\textcolor{red}{\textbf{TODO\IfNoValueTF{#1}{}{(#1)}:} #2}}
\NewDocumentCommand{\note}{o m}{\textcolor{orange}{\textbf{NOTE\IfNoValueTF{#1}{}{(#1)}:} #2}}
\newlist{todolist}{itemize}{2} \setlist[todolist]{label=$\square$}

\NewDocumentCommand{\yun}{o m}{\textcolor{blue}{\textbf{Yun\IfNoValueTF{#1}{}{(#1)}:} #2}}

\newcommand{\rev}[1]{#1}  
\newcommand{\rem}[1]{}  

\DeclareMathOperator*{\argmax}{arg\,max}
\DeclareMathOperator*{\argmin}{arg\,min}
\newcommand{\p}{\mathbb{P}} 

\newcommand{\calT}{{\cal T}}
\newcommand{\calE}{{\cal E}}
\newcommand{\MT}{{\mathbf{T}}}
\newcommand{\MM}{{\mathbf{M}}}
\newcommand{\diag}[1]{\mathrm{diag}\left(#1\right)}

\newcommand{\etal}{\textit{et~al.~}}
\newcommand{\parsec}[1]{\noindent \textbf{#1}\xspace}
\newcommand{\myParagraph}[1]{\noindent {\bf #1}\xspace}

\newcommand{\problem}{SMS\xspace}

\begin{document}

\begin{minipage}{\textwidth} This paper has been published in the proceedings of \textit{Robotics: Science and Systems}, 2024.\\ Please cite this paper as:\\
\begin{verbatim} 
@inproceedings{
  Schmid2024Khronos, 
  title = "Khronos: A Unified Approach for Spatio-Temporal Metric-Semantic SLAM
           in Dynamic Environments",
  author = {Lukas Schmid and Marcus Abate and Yun Chang and Luca Carlone},
  year = {2024},
  booktitle = "Proc. of Robotics: Science and Systems",
  year = 2024; 
} \end{verbatim} \end{minipage}

\title{Khronos: A Unified Approach for Spatio-Temporal Metric-Semantic SLAM in Dynamic Environments} 

\author{Lukas Schmid$^1$, Marcus Abate$^1$, Yun Chang$^1$, and Luca Carlone$^1$
\thanks{$^1$MIT SPARK Lab, Massachusetts Institute of Technology, MA, USA. {\tt\footnotesize \{lschmid, mabate, yunchang, lcarlone\}@mit.edu}}%
\thanks{This work is partially funded by the Amazon Science Hub “Next-Generation Spatial AI for Human-Centric Robotics” project, the ARL DCIST program, the ONR RAPID program, and the Swiss National Science Foundation (SNSF) grant No. 214489.}%
}%

\makeatletter
\let\@oldmaketitle\@maketitle
\renewcommand{\@maketitle}{\@oldmaketitle
    \centering
    \captionsetup{type=figure}
    \setcounter{figure}{0}
    \includegraphics[trim={0cm 0cm 0cm 0cm}, clip, width=\textwidth]{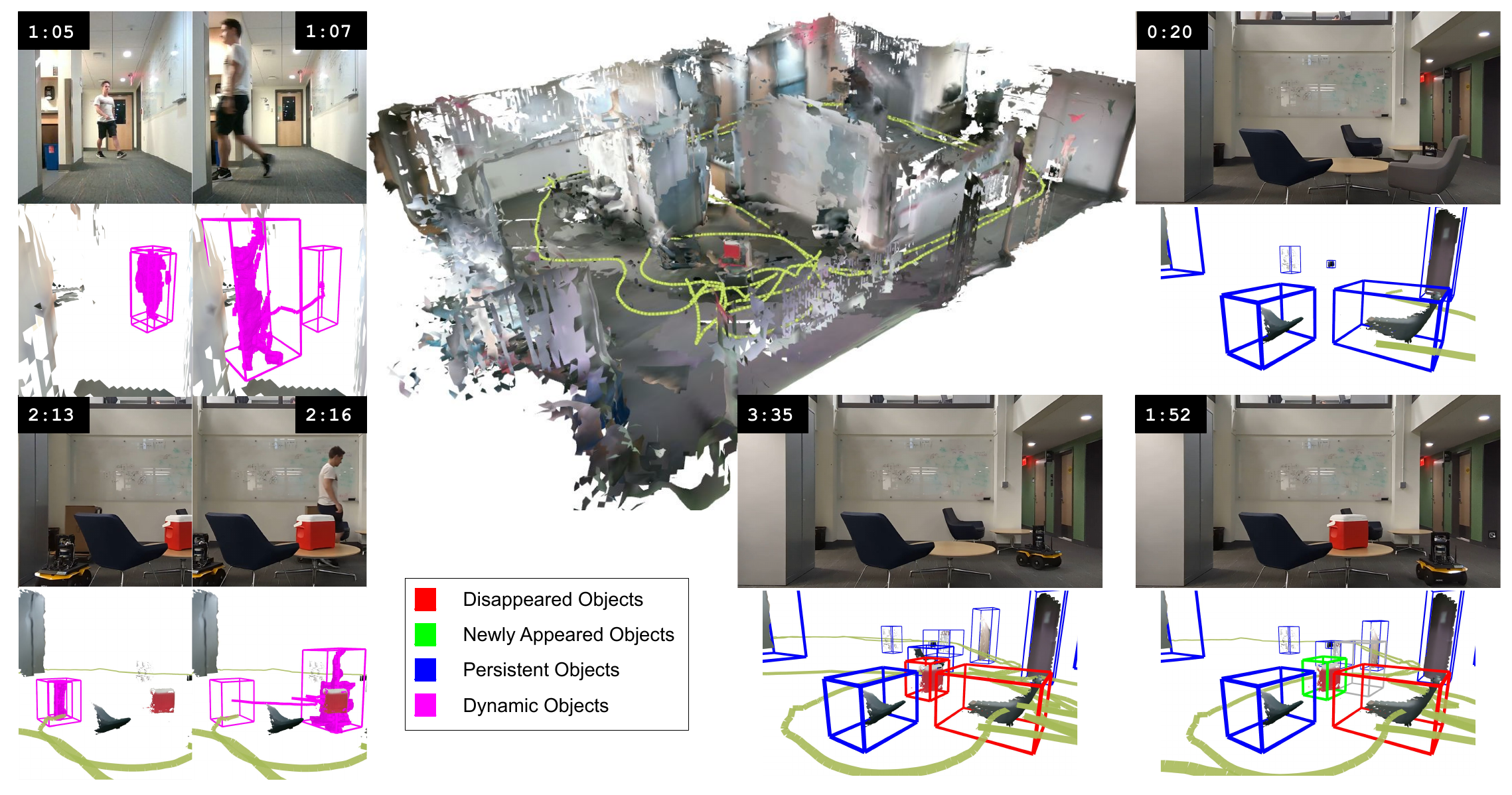}
    \vspace{-18pt}
    \captionof{figure}{
    We propose {\bf Khronos}, a unified approach to reason about short-term dynamics and long-term changes when performing metric-semantic simultaneous mapping and localization (SLAM) in dynamic environments.
    A few instances from Khronos' \emph{spatio-temporal} map, representing the scene state at all times, are shown above.
    \textbf{Short-term dynamics} (left) are shown in magenta and compared against observed human actions over the corresponding time interval. 
    \rev{We show the current and initial bounding box around the detected moving points as well as the centroid trajectory.}
    Both humans and inanimate objects \rev{such as a cart (bottom left)} are detected.
    \textbf{Long-term changes} (right) are shown for three time instances of the same scene. 
    The earliest instance is at time 0:20 (top right). 
    While the robot is moving through the hallways, a chair is removed and a red cooler is placed on top of the table; these changes are detected as the robot revisits and closes the loop at time 1:52 (bottom right).
    Lastly, the cooler is removed again, which is detected by the robot at time 3:35.}
    \vspace{-7pt}
    \label{fig:mezzanine_qualitative}
}
\makeatother

\maketitle

\begin{abstract}
Perceiving and understanding highly dynamic and changing environments is a crucial capability for robot autonomy.
While large strides have been made towards developing dynamic SLAM approaches that estimate the robot pose accurately, a lesser emphasis has been put on the construction of dense \emph{spatio-temporal} representations of the robot environment.
A detailed understanding of the \rev{scene and its evolution through time} is crucial for \rev{long-term robot autonomy and essential to} tasks that require long-term reasoning, such as operating effectively in environments shared with humans and other agents and thus are subject to short and long-term dynamics.
To address this challenge, this work defines the \emph{Spatio-temporal Metric-semantic SLAM} (\problem) problem, and presents a framework to factorize and solve it efficiently.
We show that the proposed factorization suggests a natural organization of a spatio-temporal perception system,
where a fast process tracks short-term dynamics in an active temporal window, while a slower process reasons over long-term changes in the environment using a factor graph formulation.
We provide an efficient implementation of the proposed spatio-temporal perception approach, that we call \emph{Khronos}, 
and show that it unifies exiting interpretations of short-term and long-term dynamics and is able to 
construct a \rev{dense} spatio-temporal map in real-time.
We provide simulated and real results, showing that  the spatio-temporal maps built by Khronos are an accurate reflection of a 3D scene over time and that Khronos outperforms baselines across multiple metrics. We further validate our approach on two heterogeneous robots in challenging, large-scale real-world environments.
\end{abstract}



\section{Introduction}
\label{sec:introduction}
In order to operate safely and effectively in human-populated environments, a robot needs to have a sufficient understanding of the world around it. 
Such shared spaces are often highly dynamic, with people, robots, and other entities constantly moving, interacting, and modifying the scene.
For a robot to operate in such circumstances, it is not sufficient to build a world model just for a single snapshot in time.
Instead, the robot should be also able to reason over the state of the scene at past times, inferring how the scene might have changed across multiple observations.
Such capabilities are essential for a variety of applications that require reasoning over longer time spans, ranging from household and service robotics, to industrial construction or work-site monitoring, where robots are not only required to operate in highly dynamic environments,but also to keep track of --- or reason about --- the evolution of the environment from the past to more intelligently carry-out tasks efficiently.

Metric-semantic simultaneous localization and mapping (SLAM)~\cite{Rosinol20icra-Kimera} allows a robot to construct a semantically annotated geometric representation of a scene in real-time.
Geometric information is critical for robots to navigate safely and to manipulate objects, while semantic information provides the understanding for a robot to execute human instructions and to provide humans with models of the environment that are easy to understand.
In order to build these dense metric-semantic representations in real-time, it is common to assume that the world is static and focus on robustly fusing noisy geometric and semantic measurements into a metric-semantic model~\cite{McCormac183dv-fusion++,Krishna23arxiv-3dsslam}.
Even though this is a valid assumption for certain robotic applications, it limits the generality of the types of environments a robot can operate in, along with the tasks it can be assigned to carry out.
On the other hand, there exists an extensive body of work addressing SLAM in dynamic environments~\cite{Qiu22icra-AirDOS,Cui19access-SOFSLAM,Brasch18iros-dynamicSLAM,Yu18iros-DSSLAM,Song22ral-DynaVINS,Yu21vrst-fusingSemanticObject}.
These approaches show impressive capabilities in improving robot localization and state estimation in spite of moving entities within view of the robot.
However, these prior works mostly focus on \emph{short-term} dynamics, such as people or objects currently moving in front of the camera, and the corresponding literature has often been disconnected from the body of work focusing on \emph{long-term} change detection~\cite{Langer20iros-robustObjectChangeDetection, Park21iros-ChangeSim,Schmid22icra-panopticMultiTSDF,Qian23rss-POVSLAM}, where the scene undergoes substantial changes (e.g., furniture being rearranged) while the robot is not directly observing it.
Real environments undergo both short-term and long-term changes and the literature currently lacks a unifying approach that can reason over both. Moreover, to enable robots to effectively work alongside other humans and robots, such a framework needs to  build this understanding of the world online during robot operation with the limited information and computational resources available.

To this end, we introduce the \emph{Spatio-temporal Metric-semantic SLAM} (\problem) problem, which aims at building a dense metric-semantic model of the world at all times incrementally as the robot navigates the scene.
We present a unified framework to tackle the \problem problem. 
The central idea of our approach is to develop a new factorization of the \problem problem based on spatio-temporal local consistency, which allows for the disentanglement of errors arising from sensing noise, state estimation errors, dynamic objects, and long-term changes in the scene.
We integrate this insight into a spatio-temporal perception system, named  \emph{Khronos}, which is the first real-time metric-semantic  system capable of building a spatio-temporal map of the scene.
We thoroughly evaluate our method in several simulated scenes with detailed annotations on background reconstruction, object detection, motion tracking, and change detection and on multiple robotic platforms navigating highly dynamic real-world environments. 
We make the following contributions:
\begin{itemize}
    \item We formalize the Spatio-temporal Metric-semantic SLAM (\problem) problem, which allows a robot to build a dense metric-semantic understanding of the surrounding environment and its evolution over time. 
    \item We propose a novel factorization of the \problem problem, which provides a unifying lens for existing interpretations focusing on short-term and long-term dynamics.
    \item We present Khronos, the first spatio-temporal metric-semantic perception system, composed of novel algorithms for asynchronous local mapping and deformable global change detection.
\end{itemize}

We release our implementation and datasets open-source.\footnote{Released upon acceptance at \url{https://github.com/MIT-SPARK/Khronos}.}


\section{Related Works} 
\label{sec:rel_work}

\parsec{Metric-semantic SLAM.}
The goal of metric-semantic SLAM is to build a semantically annotated 3D map during online robot operation.
Prominently, voxel-based methods~\cite{Song23tgrs-VoxelNextFusion,Li22arxiv-voxel,Shi22tcsvt-RGBDSemantic} can incrementally fuse noisy detections into the map. 
However, their rigid grid-structure makes them susceptible to state estimation drift. 
Alternatively, object-level SLAM methods~\cite{McCormac183dv-fusion++,Krishna23arxiv-3dsslam} and the related landmark-based methods~\cite{Bowman17icra-probabilisticDataAssocSemanticSLAM,Michal22iros-semanticSlamAtScale} refine the state estimate and maintain semantic information about objects in the scene. 
To achieve global consistency of the dense map, other representations such as surfels~\cite{Chen19iros-suma++}, meshes~\cite{Shi22tcsvt-meshnetSP}, or submap-based methods~\cite{Zhan21access-submapSLAM,Chen18iros-submapVSlam} have been proposed. 
Recently, also neural representations including NeRF~\cite{Maggio23icra-LocNerf,Rosinol23iros-nerfSLAM} and Gaussian-Splatting-based mapping methods~\cite{Kerbl23Ttog_GaussianSplatting,Matsuki23arxiv-GaussianSplattingSLAM,Yan23arxiv-GSSLAM} have been proposed.
However, these methods oftentimes require powerful computers, making them unsuitable for real-time robot operation.
To jointly optimize for the robot poses and semantic reconstruction of the scene, the methods above generally make the simplifying assumption that the world is static.

\parsec{Dynamic SLAM.}
To address this limitation, a large body of work has emerged, which can be grouped into two main categories.
First, sparse SLAM methods focus on improving state estimation performance~\cite{Qiu22icra-AirDOS,Cui19access-SOFSLAM,Brasch18iros-dynamicSLAM,Yu18iros-DSSLAM,Song22ral-DynaVINS,Yu21vrst-fusingSemanticObject,bescos2018dynaslam} --- typically by removing dynamic objects from the SLAM problem~\cite{Yu18iros-DSSLAM,Song22ral-DynaVINS,Yu21vrst-fusingSemanticObject}.
Alternatively, dynamic entities such as cars~\cite{Bescs20ral-DynaSLAM2,Henein20icra-DynamicSLAM} or humans~\cite{Henning22eccv-BodySLAM} can be integrated into the estimation problem, leading to improved state estimates if good motion priors for these entities are available.
Second, simultaneous tracking and reconstruction approaches~\cite{Runz17icra-cofusion,Long21ral-RigidFusion,Strecke19iccv-EMFusion,Xu19icra-midFusion,Hachiuma19arxiv-detectfusion,Xu22iros-LearningCompleteShapes} are able to generate dense models of single~\cite{Runz17icra-cofusion,Long21ral-RigidFusion} or multiple~\cite{Strecke19iccv-EMFusion,Xu19icra-midFusion,Hachiuma19arxiv-detectfusion,Xu22iros-LearningCompleteShapes}, rigid objects moving in front of the camera.
However, these approaches are often limited to table-top scenes.
Dynablox~\cite{Schmid2023ral-Dynablox} and similar approaches~\cite{Ren22iros-viMid,Mersch23ral-volumetricBeliefsDynamic} can simultaneously perform motion detection and dense background reconstruction in larger scenes. 
Nonetheless, these methods often rely on tracking of incremental motion against a static background and may not generalize well to long-term changes in the scene. 

\parsec{Change Detection.}
Identifying long-term changes is traditionally addressed in a multi-session scenario, where two observations of the static scene before and after changes occurred are compared~\cite{Langer20iros-robustObjectChangeDetection, Park21iros-ChangeSim,Zhu23arxiv-livingScenes,Sun23arxiv-NSS}.
\rev{Notably, Bore~\etal\cite{Bore19tro-generalMovableObjects} detect and also track general long-term dynamic objects using a filtered probabilistic motion model. }
Recently, the first approaches \cite{Schmid22icra-panopticMultiTSDF} for online long-term consistent mapping have emerged.
Panoptic Mapping~\cite{Schmid22icra-panopticMultiTSDF} leverages semantic consistency of foreground objects and background classes to maintain a volumetric map in changing scenes.
Fu~\etal\ extend this idea with local registration and volumetric object descriptors~\cite{Fu22ral-PlaneSDF}, as well as neural object models~\cite{Fu22iros-RobustChangeDetection}. 
Recently, POCD~\cite{Qian22rss-POCD} and POV-SLAM~\cite{Qian23rss-POVSLAM} approach the change-detection problem in a SLAM scheme and propose an object-aware SLAM pipeline to track and reconstruct object-level long-term changes in a factor-graph formulation. 

\parsec{Spatio-temporal Mapping.}
To the best of our knowledge, there are comparatively fewer works that combine short-term dynamics and long-term dynamics.
\rev{Recently,} Soares~\etal\cite{Soares23jirs-ChangingSLAM} showcase Changing-SLAM, which increases state-estimation robustness by accounting for both short-term and long-term dynamics in the scene. 
\rev{They build a map of sparse key points and employ a} Bayesian filtering approach \rev{to detect} short-term dynamics and data association \rev{to detect long-term} changes. 
\rev{Any detected} dynamic points are then removed from the SLAM problem. 
While \rev{this accounts} for \rev{the effect of} both short and long-term dynamics \rev{on} state estimation, \rev{they do not focus on} building a rich representation of the scene and an understanding of its evolution. 
In contrast, to our knowledge, our proposed method is the first to generate a spatio-temporal metric-semantic map in real-time, jointly optimizing for the robot poses and an explicit dense semantic representation of the scene at all times.
\vspace{-1pt}
\section{Problem Statement} 
\label{sec:problem}
\vspace{-1pt}

To define the spatio-temporal metric-semantic SLAM (\problem) problem, we consider a scene that is composed of objects $O_i \in \mathcal{O}$. 
The entire background of the scene is also represented as 
a static object $O_{BG} \in \mathcal{O}$. 
We denote the state of each object $i$ at time $t$ as $O_i^t$:
\begin{equation}
    O_i^t = \{ \Omega_i^t, \ T_{WO_i}^t, \ L_i \}, \label{eq:obj_attrs}
\end{equation}
where $\Omega_i^t$ denotes its surface, $T_{WO_i}^t$ its pose w.r.t. the world frame $W$, and $L_i$ the semantic label of $O_i$, at discrete time $t = 0, 1, \ldots, T$.
The background object $O_{BG} $ is constant for all times $t$.
We denote the robot poses $X$ as:
\begin{equation}
    X = \{ X^t \}^{t=0, \dots,T}, \quad X^t = T_{WR}^t.
\end{equation}
In the following, we use the shorthand notation for indexed variables of omitting the index and/or time step to refer to all existing indices and/or time steps, e.g.:
\begin{gather}
    O_i =\{ O_i^t\}^{t=0,\dots,T}, \quad O =\{ O_i^t\}_{i=0,\dots,N^t}^{t=0,\dots,T}.
\end{gather}
At each time $t$, the robot observes a volume $V^t$ which gives rise to measurements $Z_j^t$:
\begin{gather}
    Z_j^t = \{ \Omega_j^t, \ T_{RZ_j}^t, \ L_j^t \}, \\
    Z_j^t \sim O_i^t \oplus \eta_{O} \text{ if } O_i^t \in V^t .\label{eq:measurement_distrib}
\end{gather}
where $\Omega_j^t$ is a surface measurement (i.e. from an RGBD camera), $T_{RZ_j}^t$ is the pose of the object surface w.r.t. the robot frame $R$ at time $t$, and $L_j^t$ the observed semantic label of the surface at time $t$.
We summarize all measurement noise in \rev{the operation $\oplus\ \eta_O$}, which includes errors such as surface measurement inaccuracies, but also missed or hallucinated measurements, or noisy semantic information. 
Besides taking visual observations of the volume $V^t$, the robot makes an odometry measurement\rev{:}
\begin{equation}
\Phi^t \sim \left(X^{t-1}\right)^{-1}X^{t} \boxplus \eta_\Phi,
\end{equation}
\rev{where $\boxplus$ denotes addition on $SE(3)$}.
The goal of the \problem problem is to build a spatio-temporal understanding of the scene in real time, i.e.\rev{,} at each current time $T$, \rev{our goal is to} estimate the state of the scene at all previous times $t \leq T$.
This can be posed as a maximum-a-posteriori (MAP) estimation problem for each $T$:
\begin{equation}
    O^\star , X^\star = \argmax_{O, X} \p(O, X | Z, \Phi)\rev{.} \label{eq:problem}
\end{equation}
\section{Fragments and Factorization} 
\label{sec:approach}

The MAP estimate in \eqref{eq:problem} is hard to compute for a number of reasons. 
First, since the number of objects as well as their attributes can change, there is a large number of unknown variables and comparably fewer measurements.
Second, the coupling between sensing noise, imperfect state estimation, and moving and changing scenes introduces a high degree of interdependence of all variables.
This leads to having to keep all observations in memory, which causes the problem to scale poorly with space and time.
Intuitively, this reflects the challenge that disagreements between the current measurement and map belief can originate from noise in the sensing process, erroneous state estimates, objects currently moving, or the scene having changed over time.

To overcome these challenges, we propose a novel factorization of the above problem based on the key idea of \emph{spatio-temporal local consistency}.
Formally, we assume that:
\begin{gather}
   \left\lVert \left(\prod_{t'=t-\tau}^{t} \Phi^{t'} \right) \rev{\boxminus} \left( T_{WR}^{t-\tau} \right)^{-1}T_{WR}^{t} \right\rVert < \epsilon_s \ \forall t, \ \forall \tau < \delta \label{eq:spatial_consistency}, \\
     \left\lVert O^t_i \ominus O^{t-\tau}_i \right\rVert < \epsilon_t \ \forall t, \ \forall \tau < \delta .\label{eq:temporal_consistency}
\end{gather}
Intuitively, this means that both errors in state estimation \eqref{eq:spatial_consistency} and changes in the scene \eqref{eq:temporal_consistency}, while they may grow large over time, are small for short time intervals $\delta$.

\begin{figure}
    \centering
    \includegraphics[width=0.9\columnwidth]{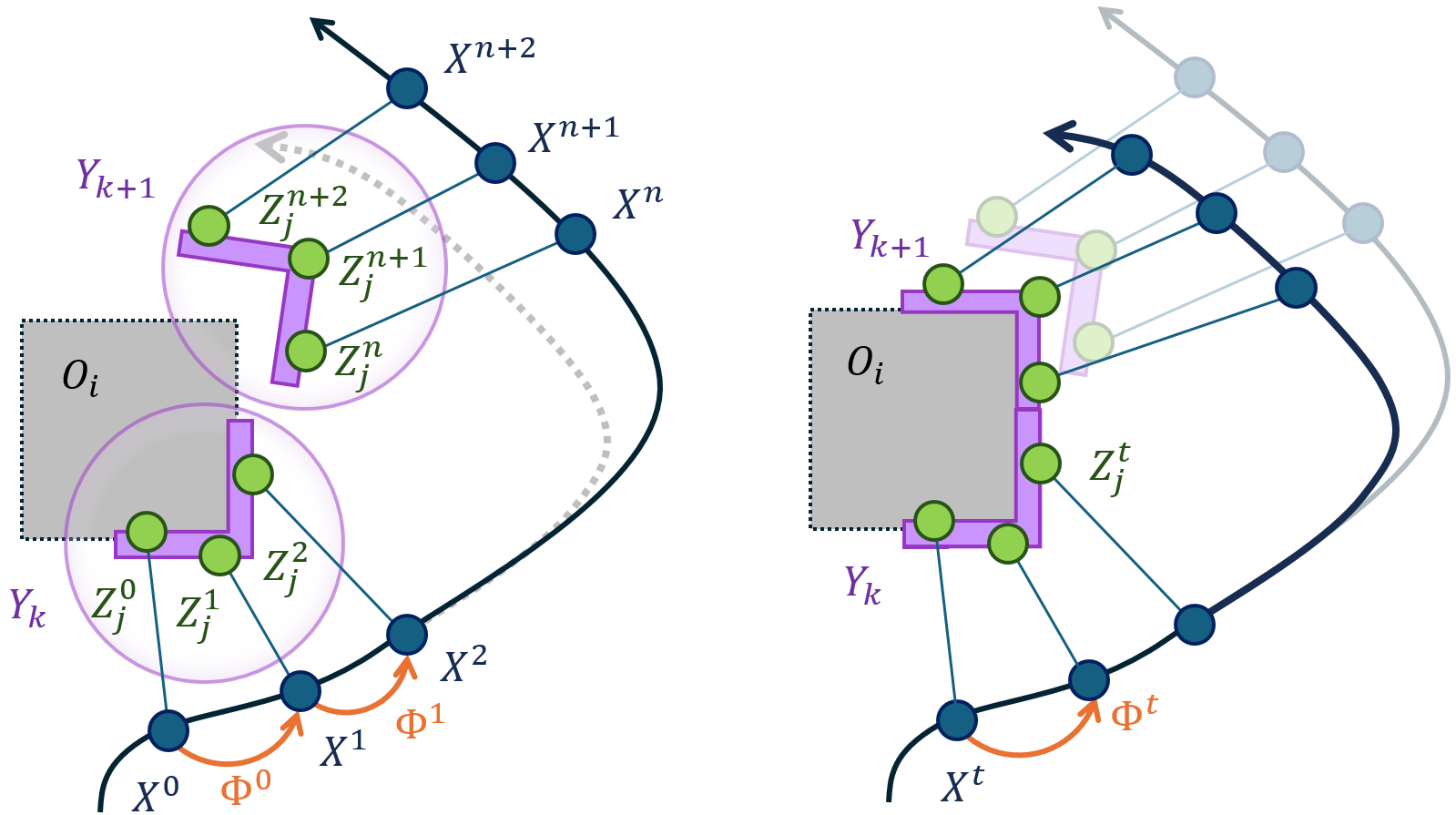}
    \vspace{-5pt}
    \caption{\rev{Key variables in our formulation. All observations $Z_j^t$ of $O_i$ are grouped into fragments $Y_k$, s.t. local consistency holds. Once it breaks, i.e., as $t(n) > t(2) + \delta$, measurements are grouped into a new fragment $Y_{k+1}$. This allows estimating $Y_k$ from the closed sets $\bar{Z}_k=\{Z_j^0,Z_j^1,Z_j^2\}$ and $\bar{\Phi}_k=\{\Phi^0,\Phi^1\}$ (left). Once the robot and fragment poses as well as associations are optimized, the object $O_i$ is fully specified by all of its fragments $\bar{Y}_i =\{ Y_k, Y_{k+1}\}$ (right).}}
    \label{fig:theory}
    \vspace{-8pt}
\end{figure}

The central idea of our approach is to decouple the dependencies in problem \eqref{eq:problem} through the introduction of a set of latent variables $Y$, which we will refer to as ``object fragments'', that serve as intermediaries between $Z$ and $O$. Each object fragment can be thought of as a partial view of an object constructed by collecting multiple locally consistent surface measurements over a short time interval (a \emph{fragment} of time).
Importantly, we define $Y$ as the minimal partitioning of the observations $Z$ of $O$ such that local consistency \eqref{eq:spatial_consistency}, \eqref{eq:temporal_consistency} holds within the observations each $Y_k$. 
Practically, this can be thought of as breaking each $O_i$ into sequences of timestamps, the fragments, where the time between consecutive observations $Z$ is $<\delta$\rev{, as illustrated in Fig.~\ref{fig:theory} (left)}.
We define properties of $Y_k^t$ \rev{in the robot frame $R$ at time $t$ as}:
\begin{equation}
    Y_k = \{ \Omega_k, \ T_{RY_k}, \ L_k \}. \label{eq:fragments}
\end{equation}
To summarize, observations $Z$ are accumulated into object fragments $Y$ based on local consistency, and each true object $O$ is a collection of fragments $Y$.
The challenge, however, is that all observations $Z$ are in robot frame $R$, whereas the goal is to estimate $O$ in $W$,
and this introduces a strong global coupling between $Z$, $X$ and $O$, as $X$ relates the $W$ and $R$ frames.
We note that the fragments $Y$ fully specify the objects $O$, hence we can express $O$ as a function of $Y$:
\begin{equation}
    \p(O, X, Y | Z, \Phi) = \p(O, X | Y, \Phi) \p(Y | Z, \Phi).
\end{equation}
Furthermore, since each $Y_k \in Y$ only depends on non-overlapping sets of measurements, we can assume that all $Y_k \in Y$ are conditionally independent, such that
\begin{equation}
    \p(O, X, Y | Z, \Phi) = \p(O, X | Y, \Phi) \prod_k \p(Y_k | Z, \Phi).
\end{equation}
We note that each $Y_k$ only depends on a subset of measurements $\bar{Z}_k \subseteq Z$, $\bar{\Phi}_k \subseteq \Phi$ that fall into its temporal window, thus yielding a first key decomposition of the \problem problem:
\begin{equation}
    \p(O, X, Y | Z, \Phi) = \underbrace{\p(O, X | Y, \Phi)}_\text{Global estimation} \prod_k \underbrace{\p(Y_k | \bar{Z}_k, \bar{\Phi}_k)}_\text{Local estimation} .
    \label{eq:problem_decomposed}
\end{equation}
This formulation isolates the effect of sensing noise in $\p(Y_k | \bar{Z}_k, \bar{\Phi}_k)$, the likelihood of a fragment $Y_k \in Y$ given measurements in the corresponding temporal window.
However, the global part of \eqref{eq:problem_decomposed} still contains a coupling between spatial and temporal sources of inconsistency that makes this a hard estimation problem.
To alleviate this, we again leverage our definition of $Y_k$ as partial observations of an object $O_i$ within a time fragment.
Since $Y$ fully specify $O$, if their poses $T_{WY_k} = X \cdot T_{RY_k}$ are known we can rewrite the global estimation part of \eqref{eq:problem_decomposed} as:
\begin{equation}
    \p(O,X|Y,\Phi) = \p(O|X,Y)\p(X|Y,\Phi).
\end{equation}
Since for a single object $O_i$, all relevant information is captured in their respective segments $\bar{Y}_i \subseteq Y$, if the associations $A_i: Y_k \mapsto \bar{Y}_i$ of fragments $Y_k$ to objects $O_i$ are known, this further simplifies to \eqref{eq:global_decomp}\rev{, as illustrated in Fig.~\ref{fig:theory} (right)}:
\begin{equation}
        \p(O,X, A|Y,\Phi) = \prod_i \p(O_i | \bar{Y}_i, X) \p(X,A |Y,\Phi). \label{eq:global_decomp}
\end{equation}
The global estimation problem is thus decomposed into two sub-problems: the fragment reconciliation term, from which the objects are inferred from their fragments, and the second term, which takes the form of a landmark-based SLAM setup.
Finally, substituting \eqref{eq:global_decomp} into \eqref{eq:problem_decomposed} yields:
\begin{multline}
     \p(O, X, Y, A | Z, \Phi) = \\
     \underbrace{\prod_i \p(O_i | \bar{Y}_i, X)}_\text{Fragment reconciliation} 
     \underbrace{\p(X, A|Y, \Phi)}_\text{SLAM}
     \underbrace{\prod_k\p(Y_k | \bar{Z}_k, \bar{\Phi}_k)}_\text{Local estimation} .
     \label{eq:problem_final}
\end{multline}
Thus, with only minimal assumptions, we have obtained a well structured problem \eqref{eq:problem_final} that we optimize as a surrogate for \eqref{eq:problem}.
More importantly, this approach provides a unified framework that naturally gives rise to existing interpretations.
First, short-term and long-term dynamics as defined in \cite{Looper2023icra-3d_vsg} naturally emerge, where \emph{all} short-term dynamics, characterized by observations of continuous motion, are captured in the local part of \eqref{eq:problem_final}, and \emph{all} long-term dynamics, characterized by observations of abrupt changes, are captured in the global part of \eqref{eq:problem_final}.
Second, the difference between modeling objects based on observations and inferring what happened while not observed is clearly represented by considering the $O_i^t$ for which $\exists Y_k^t$, as $Y_k$ group all observations of an object $O_i$, and $\nexists Y_k^t$, respectively.
Third, this formulation naturally enforces semantic consistency as introduced in \cite{Schmid22icra-panopticMultiTSDF}.
Finally, the resulting problem structure has important algorithmic properties, further detailed in Sec.~\ref{sec:system}.

\section{Khronos}
\label{sec:system}
\begin{figure}
    \centering
    \includegraphics[width=\columnwidth]{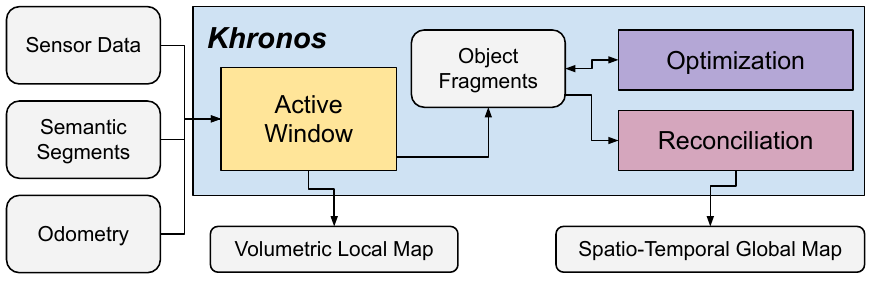}
    \vspace{-12pt}
    \caption{Khronos takes in robot odometry $\Phi$, and semantic and RGBD inputs $Z$. 
    The active window (Sec.~\ref{sec:method_local}) uses these to estimate a local representation and object fragments $Y$. 
    We then perform global optimization (Sec.~\ref{sec:method_global}) to estimate robot poses $X$ and fragment association $A$.
    Finally, the optimized fragment states are reconciled  (Sec.~\ref{sec:method_reconciliation}) to estimate the spatio-temporal map of the scene.}
    \label{fig:system}
\vspace{-10pt}
\end{figure}

This section introduces \emph{Khronos}, our system to optimize (\ref{eq:problem_final}).
We follow \eqref{eq:problem_final} and split the problem into three components, an \emph{active window}, a \emph{global optimization}, and a \emph{reconciliation} component, as shown in Fig.~\ref{fig:system}.


\subsection{Local Estimation via an Active Window}
\label{sec:method_local}
We refer to the local estimation component as  the \emph{active window}. 
Its goal is to solve the local term in \eqref{eq:problem_final} by incrementally estimating a set of fragments $Y_k$ from observations $Z, \Phi$ such that local consistency \eqref{eq:spatial_consistency}, \eqref{eq:temporal_consistency} is satisfied.

\parsec{Reconstruction.}
We first reconstruct the static background.
We point out that, while the formulation holds for any surface representation \rev{$\Omega$}, we implement Khronos using meshes to model \rev{static} surfaces \rev{$\Omega_{BG}$}.
To this end, a volumetric map is incrementally allocated around the robot and projective TSDF fusion is performed to estimate \rev{$\Omega_{BG}$}.
We then obtain candidate observations $Z$ in every frame from raw RGBD data.
Following the definition of \eqref{eq:obj_attrs}, different cues can be used to extract $Z$ from sensor data.
First, we leverage semantic masks provided in the input frame.
Second, we employ geometric motion detection to separate objects from the background.
We therefore augment the volumetric local map following the approach of \cite{Schmid2023ral-Dynablox}, \rev{using the motion cue that points falling into previously observed free space must be dynamic}.
However, since we only consider a short temporal window $\delta$ as introduced in \eqref{eq:spatial_consistency}, the \textit{Occupancy}, \textit{Sensor Sparsity} and \textit{State Estimation} terms of \cite{Schmid2023ral-Dynablox} are dropped.
\rev{In summary, we allocate one measurement $Z_j^t$ for every mask, where $\Omega_j^t$ is the set of 3D points in that mask, $T_{RZ_j}$ is the transform to the sensor, and $L_z$ is an integer or feature vector for closed or open-set semantics, respectively (see Sec.~\ref{sec:exp_semantics}).}

\parsec{Tracking.}
To estimate fragments $Y$ that best explain the observations $Z$, we generate a pool of object hypotheses $\hat{Y}$.
Since changes within the active window are small \eqref{eq:temporal_consistency}, we can greedily associate new observations \rev{$Z_j^t$} to the best fitting hypothesis \rev{$\hat{Y_k}$} by computing the volumetric IoU between each \rev{$Z_j^t$} and \rev{$\hat{Y_k}$}.
\rev{The IoU can efficiently be computed using a grid-aligned voxel-filter on the points $\Omega_k$ and $\Omega_j$.}
We associate semantic \rev{$Z_j^t$} to \rev{$\hat{Y_k}$} of \rev{matching} labels $L_k$, dynamic \rev{$Z_j^t$} to the closest dynamic \rev{$\hat{Y_k}$}, and allow cross-associations representing semantic-dynamic observations.
For each \rev{$Z_j^t$} that was not associated, a new hypothesis \rev{$\hat{Y_k}$} is added to the pool. 
Following \eqref{eq:problem_decomposed}, once local consistency is broken, no future observations can be part of the same $Y$, and these $\hat{Y}$ can be removed from the pool.
We then estimate the probability of that hypothesis $\hat{Y}$ representing a true fragment $Y$ \rev{representing an object of class $L_k$} by rejecting $\hat{Y}_k$ that have less than $\tau_{Z}=15$ observations and dynamic $\hat{Y}_k$ that have moved less than $\tau_D=\SI{1}{m}$.
We then marginalize all observations $\bar{Z}_k$ belonging to that $Y_k$.
It is worth to point out that this gives an enormous freedom in choosing suitable representations depending on the task and computation power at hand, given that all relevant observations $\bar{Z}_k$ are known at extraction time.
In this approach, we choose to use TSDF fusion to reconstruct the surfaces of static objects \rev{as meshes with adaptive resolutions $\Omega_k$}, and use \rev{sequences of} pointclouds to represent deformable \rev{and} dynamic object \rev{surfaces $\Omega_k$}.
Similarly, we track local consistency of the background and extract vertices that exit the active window. 
 This asynchronous tracking of every entity in the active window guarantees that local consistency holds and naturally handles partial or erroneous observations.
Simultaneously, it has the large advantage that object properties can be estimated when all data is available.

\subsection{Global Optimization}
\label{sec:method_global}
The global optimization module addresses the second term in \eqref{eq:problem_final};
the local estimates from the active window are optimized and updated for reconciliation.
Globally, we jointly estimate the poses of the robot $X$, the \rev{poses} of the fragments $Y$, and the dense background mesh \rev{$\Omega_{BG}$}.
In particular, we construct a deformation graph as described in~\cite{Hughes24ijrr-hydraFoundations} and augment it with fragments $Y_k$.
Specifically, the nodes of the deformation graph correspond to robot poses $X$ and mesh control points $P_M$ selected from \rev{$\Omega_{BG}$} as in~\cite{Hughes24ijrr-hydraFoundations}, and a new set of fragment poses $T_{WY_k}$.
\rev{We initialize $T_{WY_k}$ at the centroid of $\Omega_k$ with unit orientation, representing the reconstruction frame of $\Omega_k$.}
We connect $T_{WY_k}$ to the robot pose graph with edges $\calE_{\rev{XY}}$.
Each $T_{WY_k}$ is always connected to the robot poses \rev{$X_f$ and $X_l$}, corresponding to when the fragment was \emph{first}  \rev{and \emph{last} observed, respectively.
This reflects the robot-centric measurement of $Y_k$ \eqref{eq:fragments} and ensures that when the robot poses are optimized, the fragment poses are updated correspondingly.}

\begin{figure}
    \centering
    \includegraphics[width=\columnwidth]{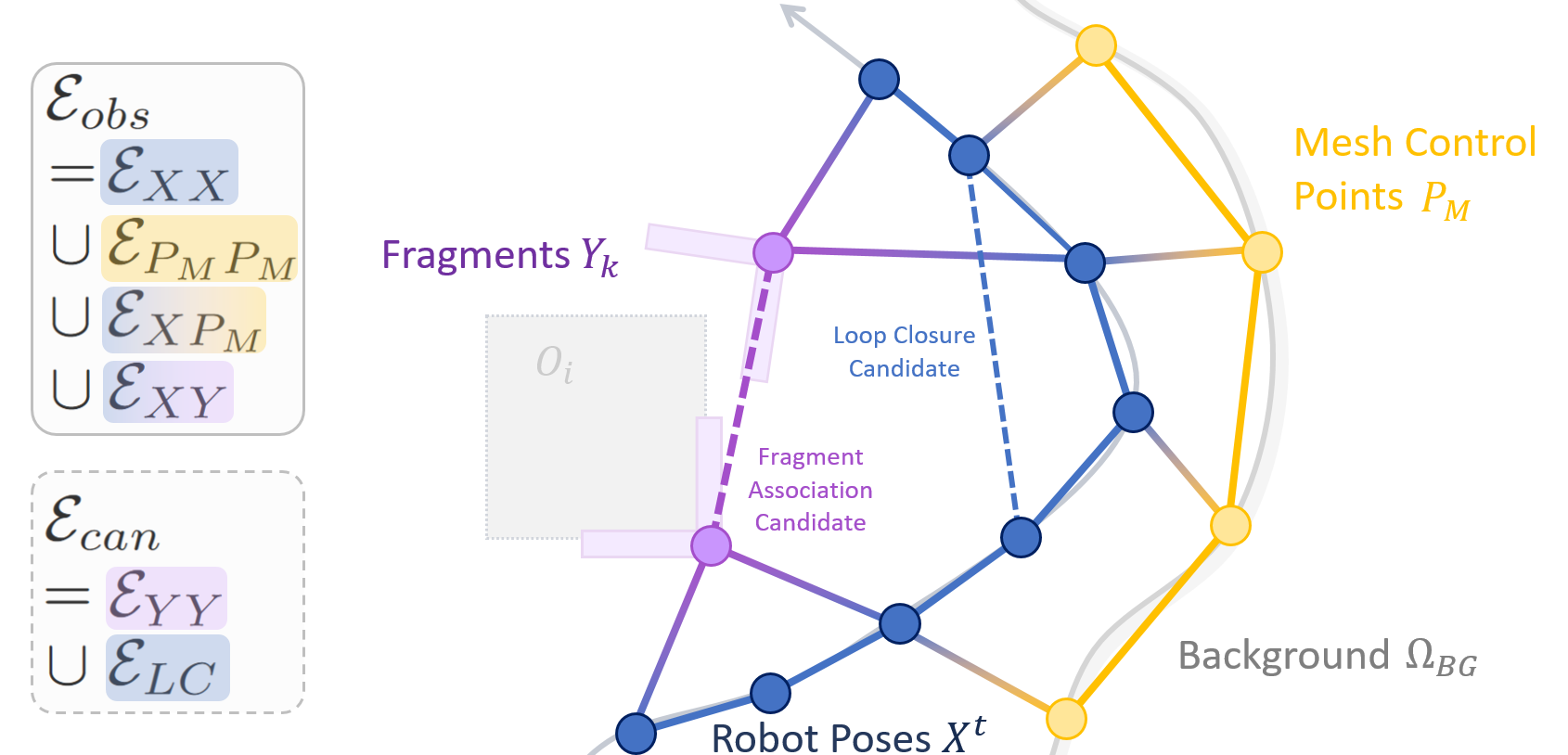}
    \caption{\rev{Overview of the constructed deformation graph $\calE$.}}
    \label{fig:dgraph}
    \vspace{-15pt}
\end{figure}

To compute associations $A$ \rev{between} fragments $Y$ when the robot re-observes an object, we \rev{generate candidate associations as edges $\calE_{YY}$ between fragments that have identical labels $L_k$ (or similar $L_k$ for open-set semantics), and whose bounding boxes overlap. 
For this initial work, we model $\calE_{YY}$ as a pure translation constraint between $T_{WY_a}$ and $T_{WY_b}$, reflecting that the centroids of $\Omega_a$ and $\Omega_b$ should be close if they stem from the same object.
We note that full 6DOF registration between $\Omega_a$ and $\Omega_b$ can naturally be integrated into our framework, but leave this for future work.
Finally, we add candidate edges $\calE_{LC}$ for loop closures provided by the odometry input.}

Hence, the edges of the final deformation graph consist of \rev{\emph{observed} edges $\calE_{obs} = \calE_{XX} \cup \calE_{P_M P_M} \cup \calE_{XP_M} \cup \calE_{XY}$, and \emph{candidate} edges $\calE_{can} = \calE_{YY} \cup \calE_{LC}$, thus $\calE = \calE_{obs} \cup \calE_{can}$}.
\rev{An overview of this is shown in Fig.~\ref{fig:dgraph}.}
The estimates of the robot poses, fragment positions, and the background mesh
is then obtained by finding a solution $\calT = X \cup P_M \cup T_{WY}$ of the \rev{robust} pose graph optimization problem:
\rev{
\vspace{-5pt}
\begin{gather}
    \calT^* = \argmin_{\substack{\MT_1, \cdots, \MT_n \in \calT \\ \omega_{ij} \in \{0 , 1 \}}} 
    \biggl[ \ \sum_{(i,j) \in \calE_{obs}} ||\MT_i^{-1} \MT_j \boxminus \bar{\MT}_{ij}||^2_{\mathbf{\Lambda}_{ij}}     \label{eq:rpgo} \\
    +\sum_{(i,j) \in \calE_{can}} \left( \omega_{ij} || \MT_i^{-1} \MT_j \boxminus \bar{\MT}_{ij}||^2_{\mathbf{\Lambda}_{ij}} + (1 - \omega_{ij})\ \bar{c}^2 \right) \biggr],
    \nonumber 
\end{gather}
}
\noindent
where $\MT_i \in SE(3)$ and $\MT_j \in SE(3)$ are pairs of 3D poses in $\calT$,
$\bar{\MT}_{ij} \in SE(3)$ is the relative measurement associated to each edge $(i,j) \in \calE$, \rev{and $\boxminus$ represents a subtraction operation on $SE(3)$ transforms.}
We use the notation $||\MM||^2_{\rev{\mathbf{\Lambda}}} = tr(\MM \rev{\mathbf{\Lambda}} \MM^T)$
and the $4 \times 4$ positive semi-definite matrix \rev{$\mathbf{\Lambda}_{ij}$} is chosen as the inverse of the covariances for the edges.
For the edges in $\calE_{YY}$, we set \rev{$\mathbf{\Lambda}_{ij}$} to $\diag{[ 0 \; 0 \; 0 \; \rev{\lambda]}}$, where the zeros cancel out the rotation residual and $\rev{\lambda}=1$ is chosen permissively since we use the centroids of the $\Omega_k$ as \rev{reference for} $\calT_{Y_k}$.

To account for erroneous measurements and perceptual aliasing, we solve the pose graph optimization problem with the Truncated Least Squares (TLS) loss, \rev{where we optimize also for binary variables $\omega_{ij}$ such that $\omega_{ij} = 1$ for correct inlier measurements and $\omega_{ij} = 0$ for outliers, where $\bar{c}$ is the outlier truncation cost.}
Using the binary weights, we can classify the correctness of the \rev{candidate} associations \rev{$\calE_{YY} \cup \calE_{LC}$}.
In practice, we solve the TLS problem with Graduated Non-Convexity (GNC)~\cite{Yang20ral-GNC} in GTSAM~\cite{gtsam}.
From the solution $\calT^*$, we directly update the robot and object fragment poses, collect the correct associations \rev{$\omega_{ij}$}, and update the dense background mesh using the mesh control points $P_M^*$ as described in~\cite{Rosinol21ijrr-Kimera}.

\vspace{-1pt}
\subsection{Reconciliation}
\vspace{-1pt}

\label{sec:method_reconciliation}
Finally, the goal of reconciliation is to estimate the state of the scene at all times $t<T$, given the optimized fragments $Y$ at time $T$.
It is important to note that fragments only contain positive observations, i.e., information about the presence of \rev{detected} objects, but not about their absence.
To resolve this evidence\rev{-}of\rev{-}absence vs. absence-of-evidence problem, we perform an additional geometric verification step.

\begin{figure}
    \centering
    \includegraphics[width=\columnwidth]{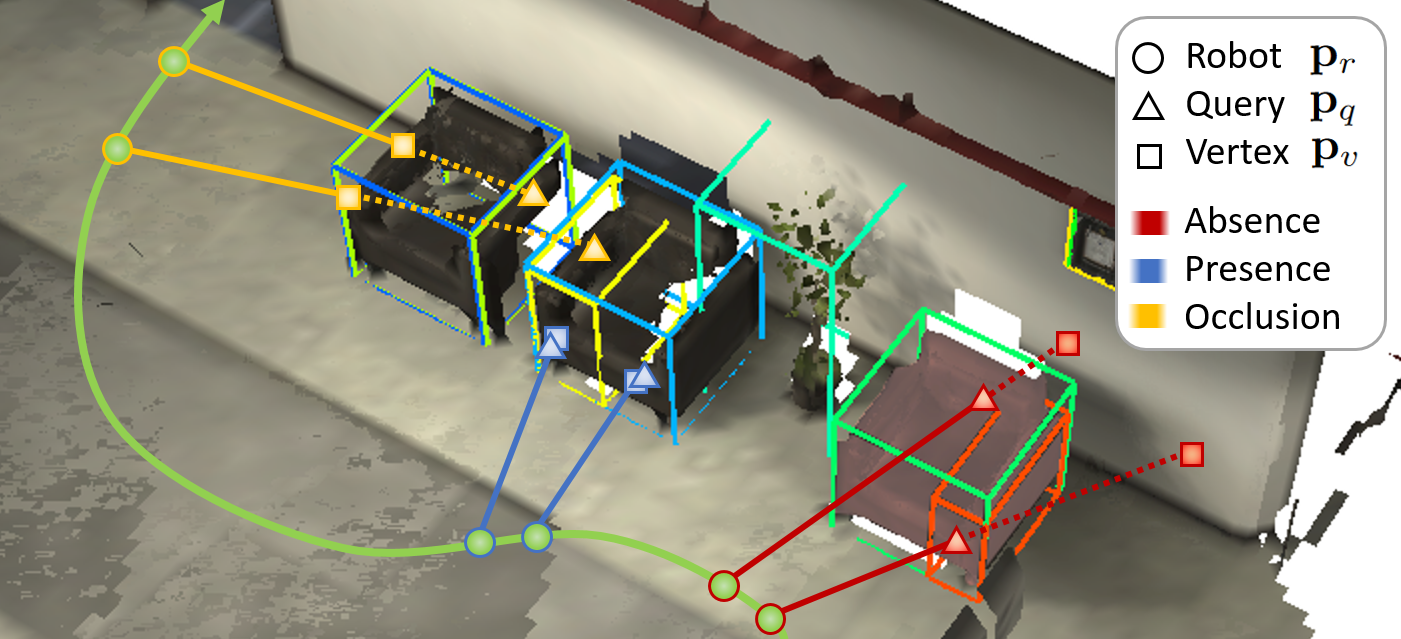}
    \caption{\rev{Deformable change detection. A $\mathbf{p}_q$ in front of $\mathbf{p}_v$ indicates the ray would pass through $\mathbf{p}_q$, and $\mathbf{p}_q$ is thus absent. Similarly, $\mathbf{p}_q$ near or behind $\mathbf{p}_v$ indicates presence or an occlusion, respectively. All $\mathbf{p}_r$ and $\mathbf{p}_v$ can freely move, rendering the approach deformable.}}
    \label{fig:change_detection}
    \vspace{-12pt}
\end{figure}

\parsec{Deformable Change Detection.}
We observe that, while no volumetric information is stored in our surface representation, this information is partially implicitly captured in the background and robot poses.
The central idea is that \rev{rays connecting a background vertex and a corresponding robot pose that observed it were in free space at that time}.
However, since the robot and background poses can continually change during global optimization, integrating this information into a free-space map is not tractable during online operation.
Instead, we approximate this global free-space information by a \emph{library of rays}.
Whenever a background vertex is extracted from the active window, we create a representative ray of that vertex \rev{$\mathbf{p}_v\in \mathbb{R}^3$} to the robot \rev{position $\mathbf{p}_r$} in the middle of its observation \rev{window}. 
For efficient lookup, this is implemented by storing the indices of the vertex and view point in a coarse global hash map.
These can efficiently and incrementally be added, and allow for the robot and background points to move freely within their grid \rev{cells}.
When large position changes are detected, e.g. in the case of a loop closure, the hash map is re-computed from scratch.
To detect evidence of absence or presence, we query points \rev{$\mathbf{p}_q$} on the surface of fragments $Y_k$ in this library of rays.
For each candidate ray returned by the hash map, we compute the \rev{distance $d_r$} of the query point to the ray to determine whether the point is on the ray\rev{, and the depth distance $d_d$ along the ray:
\vspace{-5pt}
\begin{align}
    & d_r =\ \parallel (\mathbf{p}_q - \mathbf{p}_r) \times (\mathbf{p}_r - \mathbf{p}_v) \parallel / \parallel \mathbf{p}_q - \mathbf{p}_r \parallel, \\
    & d_d = (\mathbf{p}_q - \mathbf{p}_r) \cdot (\mathbf{p}_v - \mathbf{p}_r)\ / \parallel \mathbf{p}_q - \mathbf{p}_r \parallel.
\end{align}
Distances $d_d$} longer than that of the vertex indicate the point was occluded, distances similar (within $d_{ray}=\SI{30}{cm}$) to the vertex indicate geometric consistency with that observation, and short distances indicate evidence of absence. 
\rev{An example of this logic is shown in Fig.~\ref{fig:change_detection}.}
In this way, we obtain all timestamps of rays reporting evidence of absence or presence of an object.
Since the robot poses and mesh are continuously optimized, a timestamp is only considered reliable evidence of absence if at least $c_{ray}=60\%$ percent of rays within a temporal window of $\tau_{ray}=\SI{5}{s}$ mark the object as absent.

\begin{figure}
    \centering
    \includegraphics[width=\columnwidth]{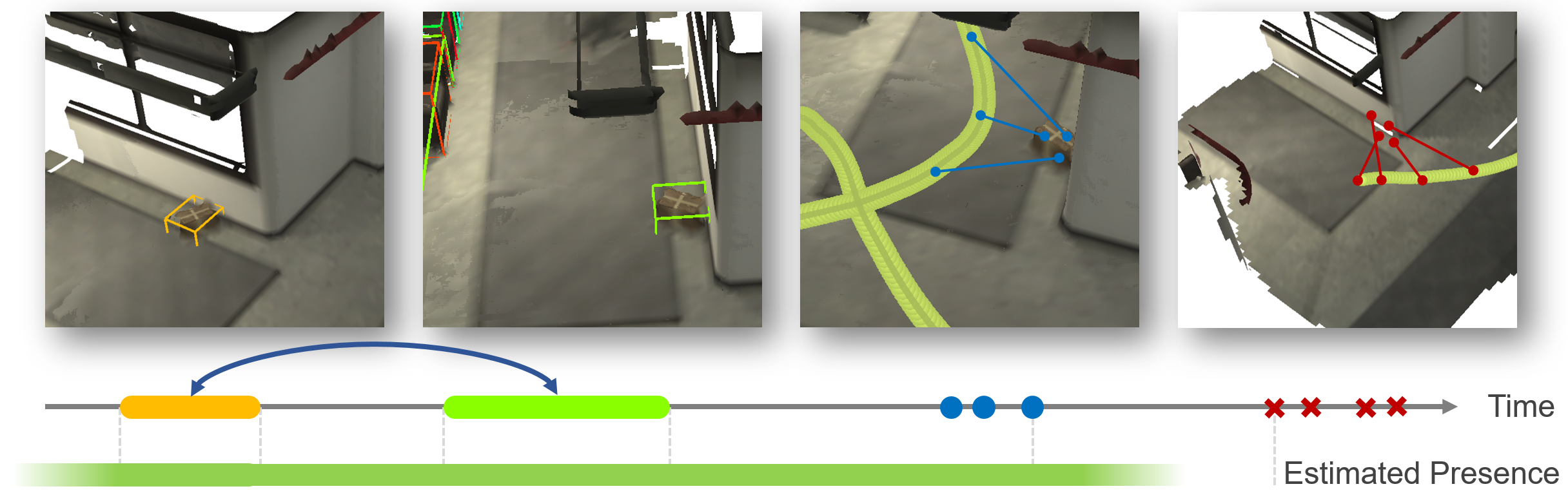}
    \caption{\rev{Reconciliation. An object is visited twice, resulting in two fragments (orange and green). Fragments are associated via global optimization (blue arrow). Even without semantic detection, geometric verification provides evidence of presence or absence. We estimate the object must have disappeared in between the last presence and first absence observation.}}
    \label{fig:reconciliation}
        \vspace{-15pt}
\end{figure}

\parsec{Reconciliation.}
Finally, we can estimate the times each object was present in between fragments $Y$.
To this end, we compute the latest absence of evidence before the fragment $Y$ was first observed and the earliest after it exited the active window. 
Similarly, we compute the earliest and latest evidence of presence within this window above. 
Intuitively, this reflects the logic that an object must have newly appeared some time between when its location was last observed to be empty and when the object was first observed in that location, and inversely for its disappearance.
Assuming that objects have a uniform probability of appearing or disappearing, the minimum expected error estimate can easily be shown to be the middle of that window.
\rev{An example is shown in Fig.~\ref{fig:reconciliation}.}

\begin{table*}[]
    \centering
    \begin{tabular}{|p{0.3em}|p{0.3em}|l|ccc|ccc|ccc|ccc|}
    \toprule
    \multirow{2}{1em}{\rotatebox[origin=c]{90}{Scene}}&\multirow{2}{1em}{\rotatebox[origin=c]{90}{Poses}}& & \multicolumn{3}{|c|}{Background} & \multicolumn{3}{|c|}{Objects} & \multicolumn{3}{|c|}{Dynamics} & \multicolumn{3}{|c|}{Changes} \\ \cmidrule{4-15}
    
    && Method & Pre & Rec & F1 & Pre & Rec & F1 & Pre & Rec & F1 & Pre & Rec & F1 \\ \midrule

   \multirow{8}{0.3em}{\rotatebox[origin=c]{90}{Apartment}} & \multirow{4}{0.3em}{\rotatebox[origin=c]{90}{GT}} 
& Hydra \cite{Hughes24ijrr-hydraFoundations}  &  94.9  & 83.1 & 87.7 & 91.0  & 28.0  & 42.3  & -  & -   & -  & -  & -   & -  \\
&& Dynablox \cite{Schmid2023ral-Dynablox} & 84.6  & \textbf{89.1} & 86.2 & - & - & - & 48.9 & \textbf{95.0} & 61.3 & - & - & - \\
&& Panoptic Mapping \cite{Schmid22icra-panopticMultiTSDF} & 95.9  & 56.1 & 70.3  & \textbf{94.6}  & 49.7  & 64.3 & - & - & - & \textbf{38.2} & 56.1  & 56.1\\
&& Khronos (ours) & \textbf{96.8}  & 87.6 & \textbf{91.2}  & 91.4  & \textbf{83.9}  &\textbf{75.3} & \textbf{90.4} & 78.6 & \textbf{84.1} & 31.3  &\textbf{69.1}  &\textbf{64.6} \\
\cmidrule{2-15}

   & \multirow{4}{0.3em}{\rotatebox[origin=c]{90}{Kimera \cite{Rosinol21ijrr-Kimera}}}
& Hydra \cite{Hughes24ijrr-hydraFoundations}  & 94.3  & 81.5 & 86.5  & 93.7  & 32.3  & 47.0  & -     & -   & - & -     & -   & -   \\
&& Dynablox \cite{Schmid2023ral-Dynablox} & 83.5  & \textbf{88.3} & 85.2 &-&-&- & 44.4 & \textbf{95.1} & 56.2 & - & - & - \\
&&Panoptic Mapping \cite{Schmid22icra-panopticMultiTSDF}  & 95.2  & 52.2 & 67.1 & \textbf{95.6}  & 46.3  & 61.8 & -     & -   & -  & \textbf{23.4}  & \textbf{65.0}  &\textbf{47.4} \\
&& Khronos (ours)  &\textbf{95.4}  & 86.7 & \textbf{90.1}  & 90.7  & \textbf{57.1} & \textbf{69.2} & \textbf{96.4} & 78.5 & \textbf{86.3} & 21.3  & 62.9 & 45.6  \\
\midrule

   \multirow{8}{0.3em}{\rotatebox[origin=c]{90}{Office}} & \multirow{4}{0.3em}{\rotatebox[origin=c]{90}{GT}} 
& Hydra \cite{Hughes24ijrr-hydraFoundations}  & 96.8  & 69.4 & 78.0  & 95.6  & 59.1  & 68.8  & -     & -   & - & -     & -   & -\\
&& Dynablox \cite{Schmid2023ral-Dynablox} & 85.9 & 73.2& 76.2 & - &-& - & 22.6 & 47.2 & 23.0 &  -&-&- \\
&& Panoptic Mapping \cite{Schmid22icra-panopticMultiTSDF}  & \textbf{97.5}  & 70.8 & 79.0  & 96.3  & 58.0  & 68.7 & -     & -   & -  & \textbf{33.1}  & 32.9  &60.4 \\
&& Khronos (ours)  &  96.3  & \textbf{73.7} & \textbf{80.6}  & \textbf{98.7}  & \textbf{63.9}  & \textbf{74.2} & \textbf{96.0} & \textbf{59.7} & \textbf{73.2} & 24.5 & \textbf{54.2}  & \textbf{66.2}  \\

\cmidrule{2-15}

   & \multirow{4}{0.3em}{\rotatebox[origin=c]{90}{Kimera \cite{Rosinol21ijrr-Kimera}}}
& Hydra \cite{Hughes24ijrr-hydraFoundations}   &83.7  & 56.8 & 65.2 & 96.9  & 49.6  & 62.1 & - & - & - & - & - & - \\
&& Dynablox \cite{Schmid2023ral-Dynablox} & 70.8  & 56.5 & 60.3 & - & - & - & 23.2 & 43.1 & 23.5 & - & - & - \\
&& Panoptic Mapping \cite{Schmid22icra-panopticMultiTSDF}  & 82.6  & 56.9 & 64.8 & 95.9  & 54.9  & 66.3 & -     & -   & -  & 9.6   & 33.5  & 38.6 \\
&& Khronos (ours) & \textbf{83.9}  & \textbf{60.3} & \textbf{67.6}  & \textbf{98.3}  & \textbf{62.4}  & \textbf{73.1} & \textbf{59.3} & \textbf{54.7} & \textbf{53.9} & \textbf{25.8}   & \textbf{52.2}  & \textbf{62.0}  \\

 \bottomrule
    \end{tabular}
    \caption{4D background reconstruction, static object, dynamic object, and change detection metrics [\%]. Higher is better for all metrics.}
    \label{tab:4dmetrics}
    \vspace{-10pt}
\end{table*}
\section{Experiments}

\subsection{Experimental Setup}
\label{sec:experiments}

\myParagraph{Datasets.}
To thoroughly evaluate the proposed method, a dataset with both short and long-term dynamics in a single sequence and detailed spatio-temporal annotations is required.
As, to the best of our knowledge, no dataset with all these features exists yet, we create two scenes using the photo-realistic simulator TESSE~\cite{Rosinol21ijrr-Kimera}. 
A smaller \emph{Apartment} dataset consists of several visits of a residential scene. 
The sequence is ~87s long, with a robot speed of up to 1m/s over a trajectory of ~39m, 64 static objects, 10 dynamic objects, and 6 long-term object changes. 
A second large-scale \emph{Office} dataset explores an extensive office scene, turning a small loop in the middle, and finally closing a large loop when revisiting some of the rooms near the start pose.
The sequence is ~217s long, with a robot speed of up to 1m/s over a trajectory of ~181m, 196 objects, 6 dynamic objects, and 8 long-term object changes. 
The dynamic objects include people and a football bouncing around.
The changing objects range from large furniture to small household items (e.g. desks, chairs, boxes, vases).
We further validate our approach in two different real-world environments on two heterogeneous robots: a Clearpath Jackal equipped with an Intel Realsense D455 RGBD camera and a Boston Dynamics Spot.
We use OneFormer~\cite{Jain23cvpr-Oneformer}, a transformer-based semantic segmentation network, to produce the semantic segments for Khronos.

\myParagraph{Metrics.}
We evaluate the ability of Khronos to reconstruct the \emph{background}, model static \emph{objects}, capture short-term \emph{dynamics}, and detect long-term \emph{changes.} 
For each category, we compute the metrics $\mathcal{L} =\{$precision, recall, F1-score$\}$.
\rev{For the background, we consider each vertex that has a corresponding ground truth (GT) vertex within 20cm a positive. For static and dynamic objects, we consider each object that has a corresponding GT object a positive. For changes, we consider each object that has the same change label (newly appeared, disappeared) as the corresponding GT object a positive.}
It is important to point out that the recall is measured with respect to all entities existing at the evaluated time $t$, and not only entities already observed by the robot.
To evaluate a spatio-temporal map \rev{$M_T(t)$}, we compute the 4D-metric over all robot \rev{times T} and belief \rev{times $t$} up to the final time \rev{$\mathcal{T}$}: 
\rev{
\begin{equation}
    \mathcal{L}_{4D} = \frac{1}{2\mathcal{T}^2} \int_0^\mathcal{T} \int_0^T \mathcal{L}\big(M_T(t)\big) dt dT
    \label{eq:metrics}.
\end{equation}
}
\parsec{Baselines.}
Since, to the best of our knowledge, there do not yet exist approaches for real-time dense metric-semantic spatio-temporal SLAM, we compare Khronos to several recent approaches specializing on the individual components.
We compare against Hydra~\cite{Hughes24ijrr-hydraFoundations}, a dense, globally-consistent metric-semantic perception pipeline, Dynablox \cite{Schmid2023ral-Dynablox}, a simultaneous dense mapping and dynamic object detection approach, and Panoptic Mapping \cite{Schmid22icra-panopticMultiTSDF}, a method for online long-term consistent volumetric mapping.

\parsec{Hardware.}
All computation is performed on an Intel i7-12700H laptop CPU with 32GB of RAM, allowing online deployment on autonomous mobile robots.

\subsection{Spatio-temporal Metric-semantic SLAM}
Tab.~\ref{tab:4dmetrics} presents detailed quantitative results for Khronos and the established baselines across both simulated datasets.
For fair comparison, all methods use identical resolutions $\nu=\SI{8}{cm}$ and sensing range of $r=\SI{5}{m}$.
To allow accurate evaluation of the systemic components, ground truth semantics from the simulator are employed (we discuss the importance of semantics in Sec.~\ref{sec:exp_semantics}). 
Finally, to assess the importance of spatial-consistency, we run each method with ground truth (GT) poses and visual-inertial-odometry estimates obtained from Kimera \cite{Rosinol21ijrr-Kimera}.


\myParagraph{Background Reconstruction.}
We observe that all methods generally achieve comparable results in terms of background reconstruction.
A slight exception is Dynablox, which does not extract static object from the background, resulting in higher recall at a lower precision.
More notably, Panoptic Mapping falsely detects previous segments of background as changed and removes them when revisiting them in the apartment scene, leading to low recall.
The importance of spatial optimization becomes apparent when looking at the office scene with drift, where Khronos and Hydra are the best performing methods. 
The difference is not as strongly reflected in the numbers, as all methods only know the drifting odometry estimate for a long time before Khronos and Hydra can close the loop toward the end of the sequence.

\myParagraph{Object Detection.}
As expected, the precision of detected objects is high given the true semantic segmentation as input. 
However, it is not 100\% as all approaches merge and filter the input detections.
However, a notable difference is apparent in recall.
Since Hydra segments objects from the background, it can not capture small objects in the scene, such as the many fine items in the apartment.
On the other, the two multi-resolution approaches \rev{Khronos and Panoptic Mapping} achieve significantly higher recall \rev{in this scene}, where the multi-hypothesis tracking and reconstruction approach of Khronos shows superior performance.

\myParagraph{Dynamic Objects.}
Dynablox shows to be highly sensitive and achieves strong recall in the apartment scene.
However, it also detects a number of false positives in this complex environment.
On the other, Khronos is able to diminish the effect of noisy measurements through its model of persistent dynamic objects.
It is apparent that the office poses a more challenging scene, where both Khronos and Dynablox detect more false positives.
The reduced recall can be explained in part by the nature of some of the observed motions, such as humans walking closely in front and away from the robot, thus preventing the free-space map from being built and the motion to be detected.
This is a fundamental limitation of the employed motion detection cue.

\myParagraph{Change Detection.}
Observing the change detection performance, we notice that online detection of changing objects in cluttered and highly dynamic scenes is a challenging problem.
Nonetheless, both methods are able to detect a significant number of changes, among some false positives.
In the small apartment scene, not many meaningful loop closures are detected, and both methods show reduced performance when operating with drifting state estimates.
However, the importance of spatial consistency for change detection becomes apparent in the large-scale office scene. 
While the precision of Panoptic Mapping breaks down for imperfect state estimates in longer sequences, Khronos is able to maintain its high performance through our joint spatio-temporal optimization and deformable change detection approach.
This demonstrates that Khronos is able to build spatio-temporal maps also of large-scale environments during online robot operation.


\subsection{Semantic Segmentation Input}
\label{sec:exp_semantics}
Khronos is able to interface with different segmentation frontends, and the construction of the spatio-temporal map is agnostic to the input semantic or instance segmentation method.
To demonstrate this, Tab.~\ref{tab:frontend_semantics} 
compares the Khronos with ground-truth semantic segmentation as the input against an 
open-set segmentation method that clusters primitive image regions given by Segment Anything~\cite{Kirillov23iccv_SegmentA} using CLIP features~\cite{Radford21icml-CLIP}, approximately based on the logic described in~\cite{Gu23icra_ConceptGraphs}.
With the different segmentation frontends, we maintain high performance compared to using ground-truth segmentation as input.
The main limitation is in terms of whether the segmentation frontend is able to accurately detect the objects that change,
The open set method suffers in recall as the "object-ness" of its detections is less well defined,
leading to false negatives on object detections.
Nonetheless, Khronos is able filter and extract meaningful objects from the noisy input observations.
These findings demonstrate that our approach can build spatio-temporal maps of various "objects" of interest.

\begin{table}
    \centering
    \resizebox{\columnwidth}{!}{%
    \begin{tabular}{|l|ccc|ccc|}
    Poses & \multicolumn{3}{|c|}{GT} & \multicolumn{3}{|c|}{Kimera} \\ \midrule
    Semantics  & Pre & Rec  & F1 & Pre & Rec    & F1  \\ \midrule
    Ground Truth & 31.3 & 69.1 & 64.6 & 21.3  & 62.9  & 45.6 \\
    OpenSet~\cite{Gu23icra_ConceptGraphs} & 44.3 & 67.2 & 64.4 & 14.6& 50.1 & 40.4 \\
    \bottomrule
    \end{tabular}}
    \caption{Change detection performance with ground truth segmentation and open-set segmentation on the Apartment sequence.}
    \label{tab:frontend_semantics}
\end{table}
\vspace{-10pt}


\begin{figure}
    \centering
    \includegraphics[width=0.8\columnwidth]{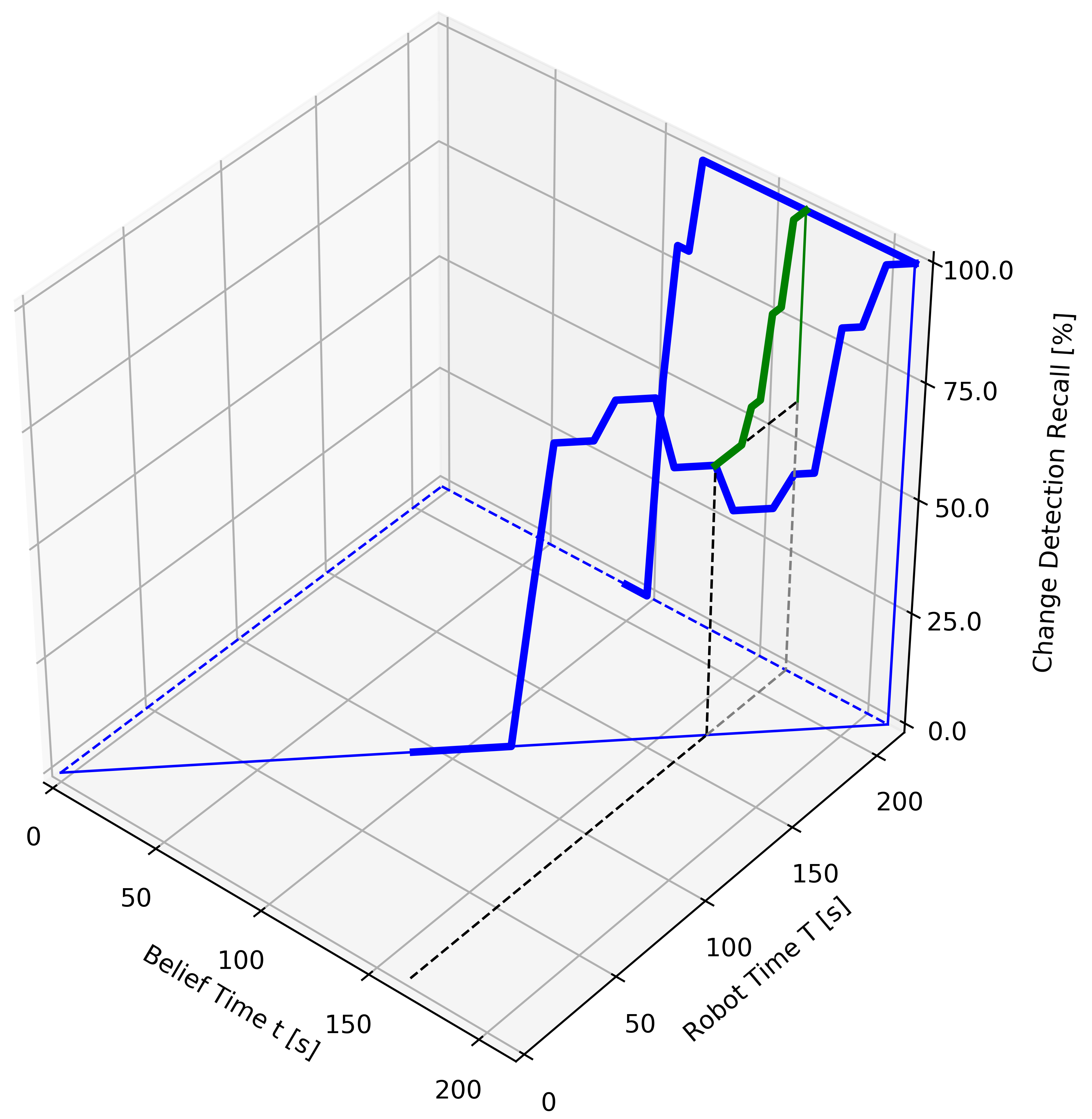}
    \caption{Spatio-temporal belief about the state of the scene at different times $t$, given observations of the robot up to its current time $T$.}
    \label{fig:4d_plots}
\vspace{-15pt}
\end{figure}

\begin{figure*}
    \centering
    \includegraphics[width=\textwidth]{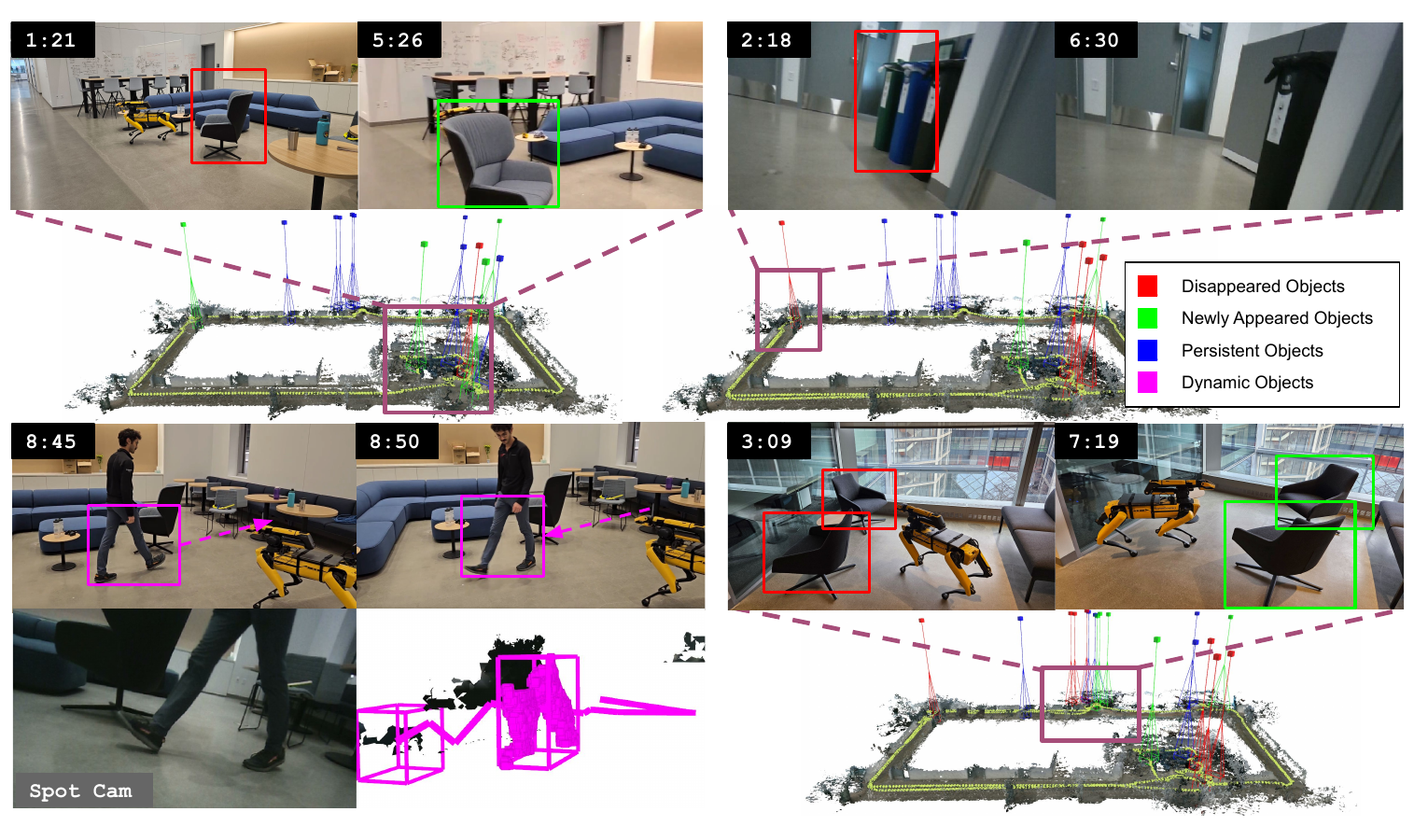} 
    \vspace{-20pt}
    \caption{The spatio-temporal map built with a Boston Dynamics Spot quadruped of an entire floor of a university building.
    We show a few instances of long term changes in the map: moving a chair to a different location (top left),
    removing two recycling bins (top right), and changing the arrangement of two chairs (bottom right).
    The Khronos reconstruction shown corresponds to the later timestamp\rev{, with the detected objects highlighted in the image.
    Note for example, how the chairs near the window (bottom right, 3:09) are present in the map at 6:30 (top right), but get marked as absent and newly appear in their new pose after the robot revisits at 7:19 (bottom right).}
    We also demonstrate short term dynamics (bottom left) with human action;
    however, due to the downward facing angle of the camera, Khronos \rev{can only see} the legs of the walking human\rev{, which are successfully detected}. }
\label{fig:spot_qualitative}
\vspace{-15pt}
\end{figure*}

\subsection{Spatio-temporal Map Beliefs}

To highlight the capability of Khronos to build spatio-teporal map beliefs, we show the change detection recall of Khronos in the office scene in Fig.~\ref{fig:4d_plots}.
One axis shows the time $T$ that has elapsed since the robot started exploring the scene. 
On the other axis, we show the belief about the world at time $t<T$, given the information obtained till $T$.
Importantly, the diagonal axis of $T=t$ represents the real-time axis, i.e., the perceived 'present' of the robot.
The first change happens around $\SI{90}{s}$, but naturally the robot has not yet observed that object again.
The robot then starts observing changes and increasing its recall, while new objects appear and disappear, decreasing the recall.
Eventually, the robot closes the loop and detects several changes at the start location.
Importantly, our approach propagates this information back through time and updates the robots belief about earlier times.
This is reflected in the belief time marginal at $T=\SI{220}{s}$, showing the robot's estimate of the scene at different times once all data is obtained.
A different way to picture this is by looking at a fixed belief time $t=\SI{165}{s}$, where we can observe how the quality of the robot belief improves as more information is gathered, highlighted in green.
Finally, the 4D-metrics \eqref{eq:metrics} can be interpreted as the area under this temporal surface.


\subsection{Mobile Robot Experiments}
We validate our approach in two real-world experiments with heterogeneous robots and sensing setups, navigating two different environments.
Fig.\ref{fig:mezzanine_qualitative} shows qualitative results of the spatio-temporal map produced by Khronos on a Jackal ground robot in a mezzanine scene.

In this experiment, the robot visits a common area (top right) and then moves around the hallway and through the kitchen. 
As it revisits the common area, a chair is removed and a cooler newly appears.
The robot then performs a longer trajectory, closes the loop back in the common area and correctly detects that the cooler has disappeared again.
In addition to long-term changes, the robot also correctly identifies people moving (left) and objects that are not semantically recognized, such as the cart being pushed through the scene (bottom left).
It is clear that Khronos is able to accurately capture both long term object appearances and disappearances, and also short term dynamic movements.
In addition, we manually annotate this smaller scale scene with all changes that were induced by the person, and present the results in Tab.~\ref{tab:real_world}.
The observed performance demonstrates that Khronos is able to accurately capture the objects in the explored scene and reflect the orchestrated changes. 
Short-term dynamic changes (human motion and a rolling cart in this case) are not reported as manual labeling was too imprecise for the dynamic objects. 

\begin{table}
    \centering
    \begin{tabular}{|ccc|ccc|ccc|}
     \multicolumn{3}{|c|}{Objects}   & \multicolumn{3}{|c|}{Changes} \\ \midrule
    Pre & Rec    & F1  & Pre & Rec    & F1 \\ \midrule
    86.9 & 97.7 & 90.2 & 87.0 & 66.7 & 93.9 \\
    \bottomrule
    \end{tabular}
    \caption{Performance of Khronos object detection and long term changes against manually annotated ground truth.}
    \label{tab:real_world}
    \vspace{-15pt}
\end{table}

Second, we perform experiments on a Boston Dynamic Spot quadruped across the entire floor of a university building.
Long-term changes to the scene are orchestrated by removing and adding objects in between robot observations.
Short-term dynamics appear in the form of humans walking or moving items in view of the robots.
Qualitative results are shown in Fig.~\ref{fig:spot_qualitative}, where we observe that Khronos scales well also to this extensive environment and a more dynamic robot platform, and is able to generate a spatio-temporal map that accurately reflects the circumstances of the scene.


\subsection{Computation Time}
\begin{figure}
    \centering
    \includegraphics[width=\columnwidth]{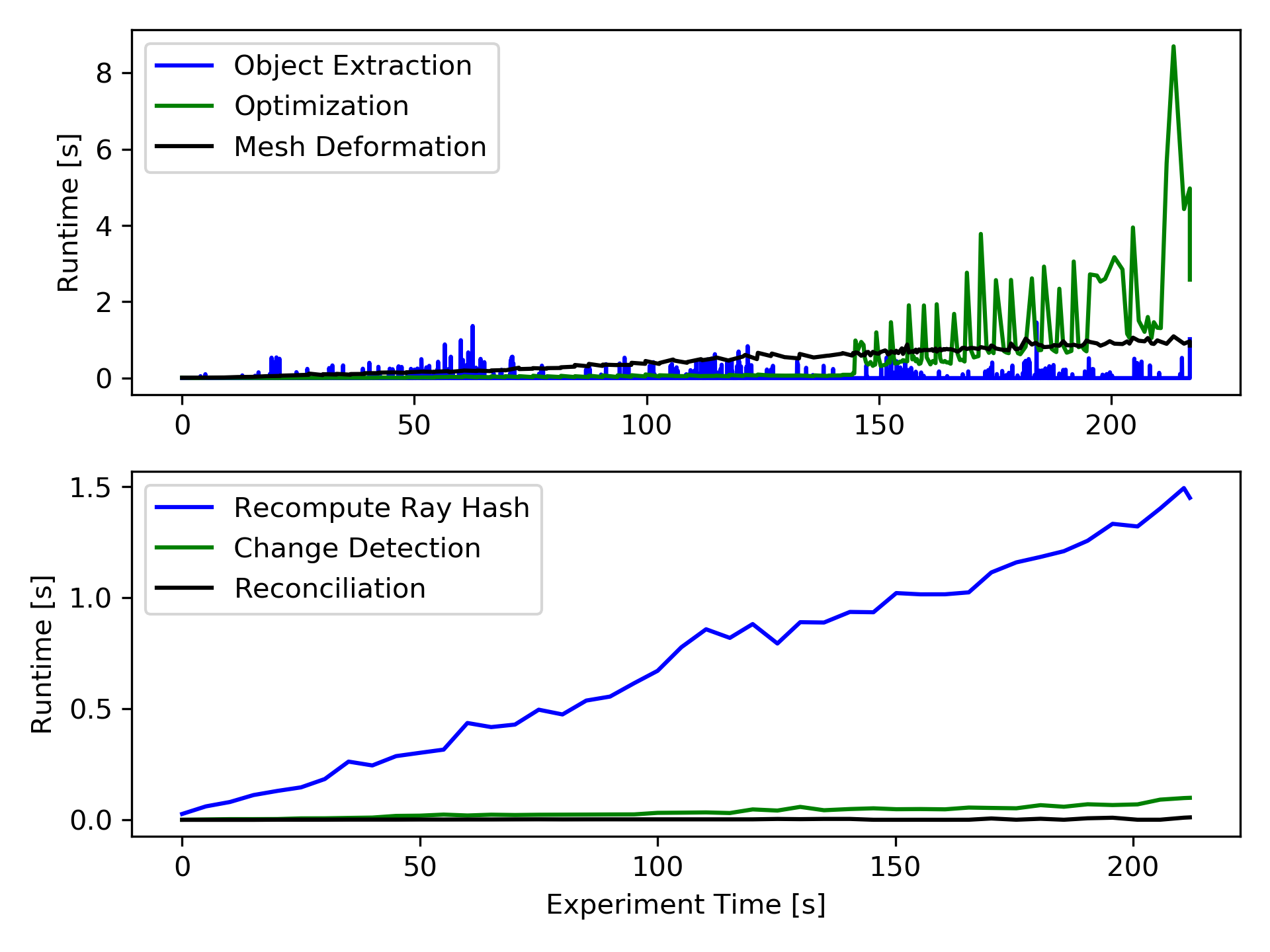} 
    \caption{Run-time of global optimization components (top) and reconciliation components (bottom).}
    \label{fig:timing}
    \vspace{-5pt}
\end{figure}

Finally, it is important for any interactive robot to perceive the scene in real-time with the limited computation available.
We therefore present timing results from the large office scene with imperfect odometry.
Due to our factorization \eqref{eq:problem_final}, the active window can operate with approximately constant time complexity.
We measure the time to process a frame in the active window at $45.5\pm9.2$ms, resulting in an average frame rate of 22.2 FPS.
The scaling of other components of our pipeline is shown in Fig.~\ref{fig:timing}.
The top row shows components of the global optimization.
The reconstruction of fragments from the active window is marked by occasional spikes, but usually takes $<\SI{1}{s}$. 
The deformation of the global background scales linearly with the observed volume, but also generally stays $<\SI{1}{s}$.
The most complex part is solving the optimization problem. 
This becomes only relevant once loop closures are detected, but requires up to several seconds in the worst case.
However, it is important to point out that all of these operations are performed in separate threads and don't need to be performed for every frame, enabling online operation of the robot.
Second, the bottom row shows timing of reconciliation components. 
Here we show worst-time performance, as the ray-hash is recomputed from scratch every time.
It is important to point out that the ray hash can also be incrementally allocated and only needs to be recomputed after a large loop closure.
Inference on the library of rays for change detection followed by reconciliation is generally quick, taking $<\SI{100}{ms}$ for all fragments in the scene.
These results demonstrate the strong algorithmic properties of the presented factorization, enabling spatio-temporal map building in real-time on mobile robots.

\section{Limitations}
\label{sec:limitations}
Since Khronos \rev{utilizes the bounding-box centroid as the position of a fragment for association edges, accurate fragment associations} can be sensitive to partial observations and occlusions.
Furthermore, the lack of \rev{6D registration between fragments} decreases the effectiveness of global estimation and reconciliation.
Adoption of modern object pose and shape estimation and registration techniques \rev{\cite{morris23arxiv-importance, judd21arxiv-MVO, Bescs20ral-DynaSLAM2}} would increase the robustness and accuracy of \rev{fragment} association.

\rev{Second, we currently only associate fragments geometrically, meaning that fragments that have moved are not associated. Incorporating fragment descriptors would allow reasoning about the history of moving objects in more detail.}

Using a ray-tracing approach for change detection implies that the existence of a reference surface.
While we found it to work well in our experiments, change detection in large open spaces with sparse surfaces would perform poorly.

Lastly, \rev{we currently keep all object fragments in memory. 
While this allows for more optimization flexibility, the problem also keeps growing indefinitely, potentially limiting the scalability of Khronos.}
In an ideal scenario, out-dated or confidently reconciled object fragments are marginalized, such that the number of object fragments stored is proportional to the number of objects to support \rev{lifelong mapping} missions.

\section{Conclusions}
\label{sec:conclusion}
In this paper, we defined the \problem problem and presented a novel approach to structure the problem, unifying the tracking of short-term dynamics and the detection of long-term changes in a single formulation
We introduced Khronos, a first metric-semantic spatio-temporal perception system capable of solving the \problem problem and generate a dense 4D spatio-temporal map.
We demonstrated that Khronos outperforms recent baselines across metrics pertaining to short and long-term dynamics, can interface with different semantic object formulations, and solve the complex \problem problem in real-time with limited compute. 
We validated Khronos on several different mobile robotic platforms, showing strong performance in complex real-world environments.

\section*{ACKNOWLEDGMENT}
We thank Jong Jin Park, Adam Fineberg, and Nathan Hughes for fruitful discussion and Nathan Hughes, Matthew Trang, and Eric Cristofalo for support with the Spot experiments.


{\small
\bibliographystyle{plainnat}
\bibliography{refs}
}


\end{document}